\documentclass{article}

\usepackage{PRIMEarxiv}

\usepackage[utf8]{inputenc} % allow utf-8 input
\usepackage[T1]{fontenc}    % use 8-bit T1 fonts
\usepackage{float}
\usepackage{hyperref}       % hyperlinks
\usepackage{url}            % simple URL typesetting
\usepackage{booktabs}       % professional-quality tables
\usepackage{amsfonts}       % blackboard math symbols
\usepackage{nicefrac}       % compact symbols for 1/2, etc.
\usepackage{microtype}      % microtypography
\usepackage{lipsum}
\usepackage{fancyhdr}       % header
\usepackage{graphicx}       % graphics
\usepackage{amsmath}
\graphicspath{{media/}}     % organize your images and other figures under media/ folder
\usepackage[table]{xcolor}
\usepackage{adjustbox}  
\usepackage[caption=false,font=normalsize,labelfont=sf,textfont=sf]{subfig}
\usepackage[square,sort,comma,numbers]{natbib}
\usepackage{algorithm,algorithmic}
\usepackage{pdflscape}

\usepackage[scaled]{helvet} % for sans-serif text
 
\usepackage{mathpazo} % for Palatino-like math fonts
\title{SyMTRS: Benchmark Multi-Task Synthetic Dataset for Depth, Domain Adaptation and Super-Resolution in Aerial Imagery}
\author{
  \href{https://orcid.org/0000-0002-5403-3911}{\includegraphics[scale=0.1]{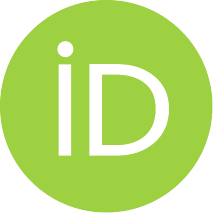}\hspace{1mm} Safouane EL GHAZOUALI*}\\
  TOELT LLC AI lab / HSLU\\
  Winterthur, Swintzerland \\
  \texttt{safouane.elghazouali@toelt.ai} \\
  \texttt{safouane.elghazouali@hslu.ai} 
  \And
  {\hspace{1mm}Nicola Venturi}\\
  Competence Center for Artificial \\ Intelligence and Simulation, armasuisse S+T, \\
  3602 Thun, Switzerland \\
  \texttt{nicola.venturi@armasuisse.ch}\\
  \And
  {\hspace{1mm}Michael Rueegsegger}\\
  Competence Center for Artificial \\ Intelligence and Simulation, armasuisse S+T, \\
  3602 Thun, Switzerland \\
  \texttt{michael.rueegsegger@armasuisse.ch}\\
  \And
  \href{https://orcid.org/0000-0002-6060-5365}{\includegraphics[scale=0.1]{orcid.pdf}\hspace{1mm}Umberto MICHELUCCI} \\
  TOELT LLC AI lab / HSLU \\
  Winterthur, Switzerland \\
  \texttt{umberto.michelucci@toelt.ai} \\
  \texttt{umberto.michelucci@hslu.ai}
}

\begin{document}
\maketitle

\begin{abstract}
Recent advances in deep learning for remote sensing rely heavily on large annotated datasets, yet acquiring high-quality ground truth for geometric, radiometric, and multi-domain tasks remains costly and often infeasible. In particular, the lack of accurate depth annotations, controlled illumination variations, and multi-scale paired imagery limits progress in monocular depth estimation, domain adaptation, and super-resolution for aerial scenes. We present \textbf{SyMTRS}, a large-scale synthetic dataset generated using a high-fidelity urban simulation pipeline. The dataset provides high-resolution RGB aerial imagery ($2048\times2048$), pixel-perfect depth maps, night-time counterparts for domain adaptation, and aligned low-resolution variants for super-resolution at $\times2$, $\times4$, and $\times8$ scales. Unlike existing remote sensing datasets that focus on a single task or modality, SyMTRS is designed as a unified multi-task benchmark enabling joint research in geometric understanding, cross-domain robustness, and resolution enhancement. We describe the dataset generation process, its statistical properties, and its positioning relative to existing benchmarks. SyMTRS aims to bridge critical gaps in remote sensing research by enabling controlled experiments with perfect geometric ground truth and consistent multi-domain supervision. The results obtained in this work can be reproduced from this Github repository: \url{https://github.com/safouaneelg/SyMTRS}.
\end{abstract}

\newpage

\section{Introduction}
Deep learning breakthroughs in remote sensing have been highly researched by large annotated datasets, still creating high-quality ground truth for diverse tasks remains a major bottleneck \cite{synrs3d2024,samrs2024}. Tasks such as geometric depth estimation, cross-domain adaptation, and super-resolution all require precise labels that are costly or infeasible to obtain at scale in real-world aerial imagery \cite{synrs3d2024,m3vir2024}. For example, monocular depth estimation in unmanned aerial vehicle (UAV) imagery requires dense 3D ground truth. Recent studies emphasize that very few real-world UAV datasets provide accurate pixel-level depth, hence why researchers rely on synthetic data or self-supervised methods that provide only relative depth maps \cite{tartanair2020,midair2019}. Synthetic simulation platforms have been used to generate aerial datasets (e.g. the Mid-Air and TartanAir datasets) with multi-modal ground truth, but this still leaves a significant sim-to-real domain gap to be addressed \cite{airsim2017}. Even when RGB images and pose or semantic labels can be readily obtained from simulations, capturing high-fidelity depth maps for aerial scenes is cumbersome and has only been achieved in a few special cases \cite{wilduav2023,usegeo2023}. 

Another fundamental challenge in remote sensing is \textit{domain adaptation} ensuring models generalize across different geographic regions, sensors, and imaging conditions. Prior high-resolution (HR) remote sensing datasets have predominantly focused on single-domain semantic mapping, overlooking issues of model transferability \cite{loveda2021}. For instance, models trained on one city often struggle to perform on another due to differences in landscape and data distribution. The LoveDA benchmark tackled one aspect of this problem by introducing a land-cover dataset with two distinct domains (urban and rural) to facilitate unsupervised domain adaptation (UDA) in semantic segmentation \cite{loveda2021}. Beyond urban-vs-rural discrepancies, \textit{illumination change} (daytime vs. nighttime) is another critical domain shift that has received little attention in aerial imaging. In autonomous driving, there have been concerted efforts to study day--night adaptation -- e.g. the Dark Zurich and NightCity datasets, and synthetic benchmarks like SHIFT \cite{shift2023} -- but in the remote sensing realm, obtaining truly co-registered day/night image pairs is extremely difficult.

High-resolution imagery is another fundamental topic in Earth observation, it delivers rich details for recognition tasks, but practical constraints like bandwidth and sensor resolution mean that many collected images are low-resolution (LR). \textit{Super-resolution (SR)} techniques aim to reconstruct finer details from LR inputs, yet training such models requires LR-HR image pairs that are representative of real-world degradations \cite{m3vir2024}. A recent effort to address this issue is the Real-RefRSSRD dataset, which provides cross-resolution pairs by pairing high-resolution aerial images (NAIP) with corresponding lower-resolution satellite images (Sentinel-2) \cite{realfused2023}. While such real-world SR benchmarks are valuable, they can be limited by temporal misalignment and differences in sensor characteristics. There remains a need for datasets that supply strictly aligned multi-scale imagery to enable controlled super-resolution experiments.

Given these limitations, we are proposing a \textit{unified multi-task datasets} that can accelerate research by providing a common testbed for multiple related problems. In computer vision at large, multi-task learning has shown promise in improving generalization and efficiency by leveraging shared representations across tasks. For example, the Taskonomy and SHIFT datasets combine diverse labels such as depth, segmentation, and tracking in one benchmark \cite{shift2023}. Another dataset, M\textsuperscript{3}VIR, was designed as a multi-modal, multi-view video benchmark containing synchronized RGB, depth, and segmentation maps \cite{m3vir2024}. In the remote sensing field, multi-task benchmark efforts are only beginning to emerge. One notable example is SynRS3D \cite{synrs3d2024}, a synthetic dataset of 69k satellite-view images providing both land-cover segmentation and pixelwise height maps. Another example is SAMRS \cite{samrs2024}, where supervised pre-training on segmentation, object detection, and change detection tasks yielded robust representation learning for remote sensing models.

In this paper, we introduce \textbf{SyMTRS}, a synthetic multi-Task dataset for transferable aerial imagery and remote sensing. SyMTRS is built upon an existing modelized urban environment named MatrixCity \cite{matrixcity}. It is constructed using a high-fidelity urban simulation in Unreal Engine 5, following the direction of recent synthetic benchmarks in vision \cite{m3vir2024} but tailored specifically to remote sensing needs. The dataset offers high-resolution RGB images ($2048\times2048$ pixels) with aligned ground-truth depth maps, paired day-time and night-time renders for controlled domain adaptation, and multi-scale image sets (downsampled $\times2$, $\times4$, $\times8$) to support super-resolution. Unlike prior benchmarks, SyMTRS ensures all annotations are spatially and temporally aligned, allowing for the study of multi-task and cross-domain models in a unified setting.

SyMTRS can be considered as a step towards holistically comprehending scenes in aerial images through domain-adaptation-based geometrically aware and resolution enhancement approaches, which also take advantage of close-coupled supervisory information. The dataset offers an opportunity for pretraining and transfer learning, provides a benchmark for individual tasks, and also enables joint objective optimization within remote sensing vision models.

\section{Related Work}
Deep learning in computer vision has benefited from large-scale annotated datasets spanning multiple tasks, particularly in ground-level imagery. In contrast, remote sensing datasets are typically narrower in scope, often addressing a single task and lacking in large-scale multi-modal supervision. In this section, we review prominent datasets from both domains---ground-level and remote sensing---and highlight the gaps that our proposed SyMTRS dataset aims to fill.

\textbf{Ground-Level Vision Datasets} \\
Ground-level datasets for classification, segmentation, and depth estimation are rich in scale and annotations. For example, ImageNet~\cite{imagenet2012} and Places365~\cite{zhou2017places} provide millions of images for object and scene classification. Datasets such as PASCAL VOC~\cite{everingham2010pascal} and MS COCO~\cite{lin2014microsoft} offer annotations for object detection and instance segmentation. Urban scene segmentation benchmarks like Cityscapes~\cite{cordts2016cityscapes} and multi-task datasets like KITTI~\cite{geiger2013vision} and NYU Depth V2~\cite{silberman2012indoor} provide pixel-level depth and segmentation labels, enabling rich multi-task learning.
Synthetic datasets such as Virtual KITTI~\cite{cabon2020virtual} and SYNTHIA~\cite{ros2016synthia} offer pixel-perfect annotations for depth, segmentation, and optical flow. The BDD100K dataset~\cite{yu2020bdd100k} combines object detection, semantic segmentation, and lane detection across driving videos, further enhancing the scope for multi-task perception.

\textbf{Remote Sensing Datasets}\\
Remote sensing datasets have historically been fragmented across tasks. For classification, UCMerced~\cite{yang2010ucmerced}, AID~\cite{xia2017aid}, and NWPU-RESISC45~\cite{cheng2017remote} provide aerial scene categories. EuroSAT~\cite{helber2019eurosat} and BigEarthNet~\cite{sumbul2019bigearthnet} offer land-cover classification at scale, but lack pixel-level supervision.
Semantic segmentation benchmarks include DeepGlobe~\cite{demir2018deepglobe}, ISPRS Potsdam~\cite{isprs_potsdam}, and LoveDA~\cite{wang2021loveda}. Object detection is addressed by DOTA~\cite{xia2018dota}, xView~\cite{lam2018xview}, and RarePlanes~\cite{rareplanes2020}, though each targets a narrow domain.
For super-resolution, recent datasets such as OLI2MSI~\cite{wei2021oli2msi} and SEN2NAIP~\cite{sen2naip2023} provide paired multi-resolution imagery. Temporal and video-based datasets, while rare, include the Jilin-1 satellite video benchmark~\cite{wang2022satvideosr} and the multi-temporal fMoW~\cite{fmow2018}. Most remote sensing datasets remain single-task and lack consistent supervision across modalities.

\textbf{Multi-Task Synthetic Benchmarks}\\
Beyond single-task collections, several recently proposed benchmarks aim to support joint supervision across multiple vision tasks by providing aligned annotations and complex scene variations.
\textbf{SynRS3D} is a large-scale synthetic remote sensing dataset developed to facilitate 3D semantic understanding from monocular high-resolution imagery. SynRS3D comprises high-resolution optical images covering diverse urban styles and multiple land cover categories, and it provides precise annotations for height (elevation) estimation, semantic land cover mapping, and building change detection. This combination enables joint training of both geometric and semantic tasks in the context of remote sensing, addressing the scarcity of ground truth data for height and semantic labels in real satellite imagery. Synthetic annotations include pixel-aligned land cover classes and height maps, enabling models to learn 3D structure and semantics simultaneously. SynRS3D also serves as a platform for exploring unsupervised domain adaptation and transfer from synthetic to real remote sensing data through multi-task baselines, helping mitigate the gap between rendered and real scenes.
\textbf{SAMRS} (Segment Anything Model Remote Sensing) extends large-scale segmentation datasets into the remote sensing domain by leveraging the Segment Anything Model (SAM) to efficiently generate pixel-level semantic annotations from existing object detection collections. The resulting SAMRS dataset contains over 105\,000 images and more than 1.6 million instance annotations, orders of magnitude larger than previous high-resolution remote sensing segmentation benchmarks. These annotations support semantic segmentation, instance segmentation, and object detection, either independently or in combination. SAMRS thereby enables pre-training and fine-tuning strategies that alleviate the annotation bottleneck in remote sensing segmentation tasks, bridging the gap between object detection and dense pixel labeling. Another multi-task dataset is \textbf{M\textsuperscript{3}VIR}, or the Multi-Modality Multi-View Synthesized Benchmark, is a recent synthetic video dataset that emphasizes multi-task support in ground-level imagery. Although efforts in this area are still emerging, M\textsuperscript{3}VIR provides multi-view imagery with precise ground truth for tasks such as depth estimation, semantic segmentation, and restoration, enabling models to learn consistent representations across views and modalities. The availability of video sequences with diverse content and aligned ground truth labels supports research into robust multi-task and multi-modal learning beyond static images, including scenarios that combine depth cues with semantic understanding and image enhancement tasks. Moreover, \textbf{SHIFT} (Synthetic Driving dataset for Continuous Multi-Task Domain Adaptation) focuses on continuous domain shifts in synthetic driving contexts and is particularly relevant for multitask evaluation under diverse environmental conditions. SHIFT includes comprehensive sensor streams with annotations for semantic segmentation, instance segmentation, monocular depth estimation, and optical flow. It simulates gradual variations in weather (cloudiness, rain, fog), time of day (day to night), and scene complexity, providing a controlled environment to study domain adaptation and continual performance degradation across tasks. By simulating these continuous domain transitions and offering dense ground truth for multiple perception tasks, SHIFT enables investigations into domain-robust multi-task models and continuous adaptation strategies.

These benchmarks illustrate recent progress toward datasets that provide dense, multi-modal annotations suitable for training and evaluating deep models on numerous vision tasks simultaneously. However, most existing multi-task collections remain grounded either in terrestrial imagery (such as driving or urban scenes) or focus on subsets of relevant tasks. Remote sensing, in particular, still lacks large-scale benchmarks that jointly support depth estimation, semantic segmentation, super-resolution, and domain adaptation which motivates the design of our proposed SyMTRS dataset.

\subsection{Comparison of Datasets}
Table~\ref{tab:dataset_comparison} provides an overview of ground-level and remote sensing datasets, summarizing their scope, resolution, modality, and supported tasks. SyMTRS is designed to bridge the identified gaps by offering a synthetic, high-resolution, multi-task benchmark for aerial vision.

\begin{landscape}
\begin{table}[p]          % [p] is often better for full-width landscape tables
\centering
\caption{Comparison of Representative Ground-Level and Remote Sensing Datasets.}
\label{tab:dataset_comparison}

\begin{tabular}{@{} l l l c c c c @{}}
\toprule
\textbf{Domain} & \textbf{Dataset} & \textbf{Tasks} & \textbf{\# Images} & \textbf{Resolution} & \textbf{Synthetic/Real} & \textbf{Single/Multi} \\
\midrule
Ground & ImageNet~\cite{imagenet2012} & Classification & 1.28M+ & Varied & Real & Single \\
Ground & Places365~\cite{zhou2017places} & Scene Classification & 1.8M+ & $256\times256$ & Real & Single \\
Ground & PASCAL VOC~\cite{everingham2010pascal} & Cls/Det/Seg & 11.5k & $500\times400$ & Real & Multi \\
Ground & MS COCO~\cite{lin2014microsoft} & Det/Seg & 118k & $640\times480$ & Real & Multi \\
Ground & Cityscapes~\cite{cordts2016cityscapes} & Segmentation & 25k & $2048\times1024$ & Real & Single \\
Ground & KITTI~\cite{geiger2013vision} & Det/Depth/Stereo & 7.5k+ & $1242\times375$ & Real & Multi \\
Ground & NYU Depth V2~\cite{silberman2012indoor} & Depth/Segmentation & 1.4k & $640\times480$ & Real & Multi \\
Ground & SYNTHIA~\cite{ros2016synthia} & Depth/Segmentation & 9.4k & $1280\times760$ & Synthetic & Multi \\
Ground & Virtual KITTI~\cite{cabon2020virtual} & Multi-task & 21k+ & $1242\times375$ & Synthetic & Multi \\
Ground & BDD100K~\cite{yu2020bdd100k} & Det/Seg/Lane & 100k & $1280\times720$ & Real & Multi \\
Ground & M\textsuperscript{3}VIR~\cite{chen2024m3vir} & Seg/Depth/SR & 100k+ & Multi-Res & Synthetic & Multi \\
\midrule
Remote & UCMerced~\cite{yang2010ucmerced} & Scene Classification & 2.1k & $256\times256$ & Real & Single \\
Remote & AID~\cite{xia2017aid} & Scene Classification & 10k & $600\times600$ & Real & Single \\
Remote & NWPU-RESISC45~\cite{cheng2017remote} & Scene Classification & 31.5k & $256\times256$ & Real & Single \\
Remote & EuroSAT~\cite{helber2019eurosat} & Classification & 27k & $64\times64$ & Real & Single \\
Remote & BigEarthNet~\cite{sumbul2019bigearthnet} & Multi-label Class & 590k+ & $120\times120$ & Real & Single \\
Remote & DeepGlobe~\cite{demir2018deepglobe} & Segmentation & 803 & $2448\times2448$ & Real & Single \\
Remote & ISPRS Potsdam~\cite{isprs_potsdam} & Segmentation & 38 & $6000\times6000$ & Real & Single \\
Remote & LoveDA~\cite{wang2021loveda} & Segmentation (DA) & 18k & $1024\times1024$ & Real & Single \\
Remote & DOTA~\cite{xia2018dota} & Detection & 2.8k & $800--4000px$ & Real & Single \\
Remote & RarePlanes~\cite{rareplanes2020} & Detection & 50k+ & $1024\times1024$ & Synthetic+Real & Single \\
Remote & OLI2MSI~\cite{wei2021oli2msi} & Super-Resolution & 5.3k & $480\times480$ & Real & Single \\
Remote & SEN2NAIP~\cite{sen2naip2023} & Super-Resolution & 38k & $600\times600$ & Real & Single \\
Remote & Jilin-1~\cite{wang2022satvideosr} & Video SR & 201 clips & $128\times128$ (video) & Real & Single \\
Remote & fMoW~\cite{fmow2018} & Temporal Classification & 1M+ & $224\times224$ & Real & Single \\
Remote & SynRS3D~\cite{song2024synrs3d} & Segmentation/Depth & 69k & High-Res & Synthetic & Multi \\
Remote & SAMRS~\cite{wang2024samrs} & Detection/Change/Seg & 100k+ & Varied & Real & Multi \\
\bottomrule
\end{tabular}
\end{table}
\end{landscape}

\section{SyMTRS Dataset}
The SyMTRS dataset is built upon the high-fidelity MatrixCity project \cite{matrixcity}, implemented within the Unreal Engine 5 simulation environment. MatrixCity is a photorealistic, procedural urban environment designed for testing computer vision, robotics, and AI algorithms in large-scale, dynamic cityscapes. Developed to leverage the graphical and physical realism of Unreal Engine 5, MatrixCity includes diverse city blocks populated with detailed building geometries, roads, vehicles, and ambient urban elements. This environment supports dynamic lighting, shadow rendering, and camera control, making it highly suitable for simulating aerial and ground-level views in a controllable setting. MatrixCity is particularly well-suited for synthetic dataset generation due to its scalability, deterministic rendering pipeline, and ability to produce multi-modal outputs such as RGB, depth, and semantic masks.

\subsection{Data Design}
The dataset was generated by deploying a customized Camera Actor within the MatrixCity environment. This camera simulates a drone-mounted sensor, configured with a physical sensor size of 35\,mm by 35\,mm and a focal length of 36\,mm to ensure accurate perspective projection and compatibility with standard camera models used in photogrammetric pipelines. No additional distortion or lens deformation effects were applied in order to preserve pixel-level geometric precision and maintain clean, artifact-free imagery.

\begin{figure}[ht]
    \centering
    \includegraphics[width=\linewidth]{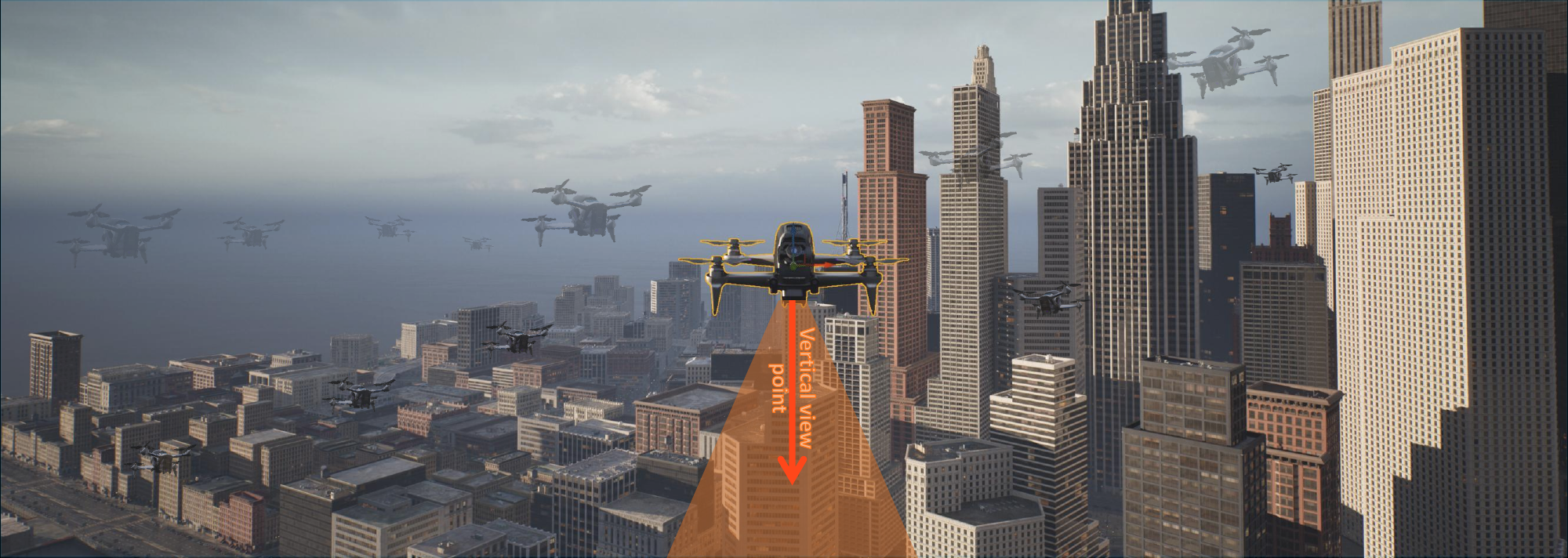}
    \caption{Visualization of the capturing process of the image in the MatrixCity Unreal Engine 5 environment.}
    \label{fig:SimulationDrone}
\end{figure}

To simulate aerial observations, the camera was oriented with a fixed pitch of $-90^\circ$, capturing nadir (top-down) views of the cityscape. The motion of the camera was automated to follow a rasterized sweeping pattern, covering the entire accessible area of the MatrixCity map. This rasterization was repeated at multiple altitudes, beginning at 90,000 Unreal Engine units and descending incrementally to 30,000 units. These altitudes simulate varying drone flight heights, offering multi-scale coverage of the urban scene.
To prevent collisions and preserve line-of-sight over the complex urban geometry, low-altitude flights were restricted from covering zones with densely packed high-rise structures. A pre-analysis of building height distribution across MatrixCity was performed, and exclusion masks were applied to regions where tall structures could occlude the camera's view or cause unrealistic overlaps during rendering.
For the illumination settings, we introduced variation in solar lighting by modifying the directional light source orientation across sequences. This variation introduces natural diversity in shading, highlights, and cast shadows throughout the dataset, avoiding bias from uniform lighting conditions. Rendering sequences were orchestrated using Unreal Engine’s built-in Sequencer tool, which enabled precise animation control and frame scheduling. Each sequence was rendered at a framerate of 60 FPS, producing temporally coherent and motion-stable image sequences.
In total, over 1.5 million frames were generated during the dataset creation phase. To avoid redundancy and ensure diverse scene coverage, a sampling step of one frame every 600 intervals was applied, yielding a balanced dataset of spatially and temporally distributed samples. This subsampling also mitigated motion blur while preserving a sufficient range of camera positions for tasks like depth estimation and domain adaptation.
Ground-truth depth maps were rendered alongside RGB images using the Movie Render Queue in Unreal Engine. The depth maps preserve real-world metric accuracy and were saved in EXR format to maintain high dynamic range and full-precision floating point values. Following rendering, the outputs were post-processed into two data streams: RGB images saved as lossless PNGs, and depth maps serialized as \texttt{*.npy} files in 32-bit float format, ensuring compatibility with scientific computing libraries and high-fidelity downstream processing.

% \textcolor{red}{[Additional metadata, such as camera intrinsics, frame indices, sun direction vectors, and geographic bounding boxes, will be included in the dataset release to facilitate reproducibility and enhance usability across a range of vision tasks.]}

\subsection{Vision Tasks}

\begin{figure}[ht]
    \centering
    \includegraphics[width=\linewidth]{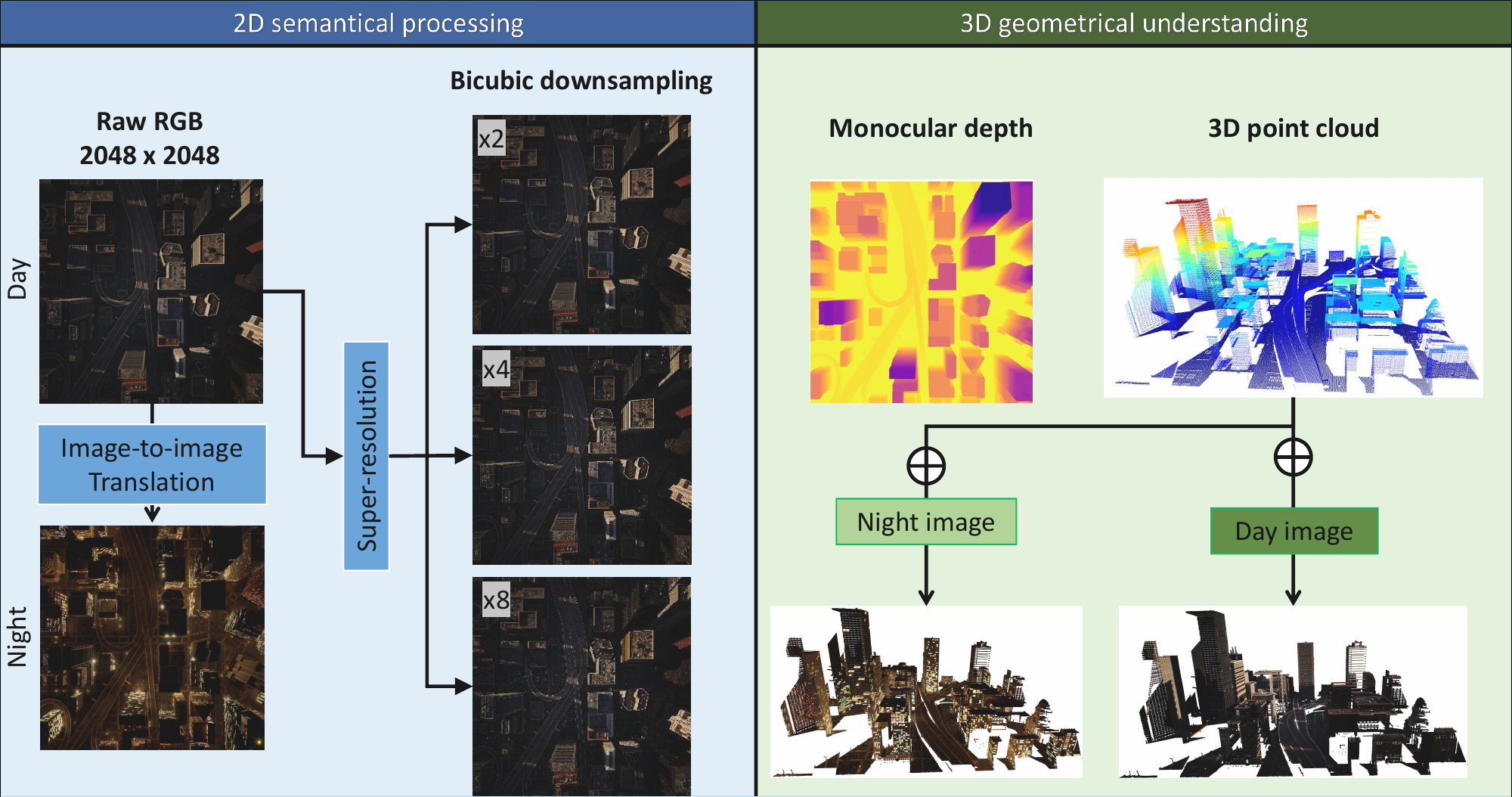}
    \caption{Sample representation of the dataset components which include: Raw RGB high resolution images $(2048\times2048)$ captured at different $Z$ heights of the map. Night version of the image with lit building blocks $(2048\times2048)$. Bicubic downsamples for super resolution at 3 scales $(1024\times1024)$, $(512\times512)$ and $(256\times256)$. and provided depth map with metric values stored in numpy array for conversion to 3D point clouds.}
    \label{fig:SyMTRS_dataset_overview}
\end{figure}

SyMTRS is designed to support multiple vision tasks as shown in Fig. \ref{fig:SyMTRS_dataset_overview}. The settings that are presented within our current benchmark protocol are: (1) \textbf{Super-resolution (SR)} paired LR$\rightarrow$HR reconstruction at $\times2$, $\times4$, and $\times8$ from perfectly aligned synthetic pairs. (2) \textbf{Image-to-image translation (paired)} day$\rightarrow$night translation with paired supervision. (3) \textbf{Unsupervised image generation} to synthesis further images. %using CycleGAN~\cite{zhu2017cyclegan}.
For the current experiments, we use a fixed split with seed $42$, containing 1656 training tiles and 414 test tiles. The current release reports SR and day/night translation experiments; depth and detection benchmarks are planned for the next release stage.

\section{Experiments}
This section reports the training setup and the first benchmark results obtained so far on SyMTRS for SR and day/night translation. All runs use a deterministic split seed ($42$). For SR, the split contains 1656 training tiles and 414 held-out test tiles. To benchmark SyMTRS dataset, we have used a Linux machine equipped with the components presented in Table \ref{tab:hardware_software}.

\begin{table}[ht]
    \centering
    \renewcommand{\arraystretch}{1.} % Adjust row height for better readability
    \caption{Computational infrastructure used during the training and evaluation of object detection models. Details on GPUs, CPU, RAM, operating system, and key software frameworks such as PyTorch, Ultralytics, and CUDA versions are described.}
    \begin{tabular}{l l}
        \hline
        \textbf{Hardware} & \textbf{Software} \\ 
        \hline
        GPUs: NVIDIA RTX A6000 48 GB × 3 & Ubuntu 20.04.6 LTS \\ 
        CPU: Intel(R) Core(TM) i9-10980XE @ 3.00GHz & Python 3.10.15 \\ 
        Memory: 128 GB & PyTorch 1.13.1 \\ 
        & Ultralytics 8.3.9 \\ 
        & CUDA 12.4 \\ 
        \hline
    \end{tabular}
    
    \label{tab:hardware_software}
\end{table}

\subsection{Super-resolution}

Single-image super-resolution (SR) aims to reconstruct a high-resolution (HR) image from a low-resolution (LR) observation, which is an inherently ill-posed inverse problem since multiple plausible HR solutions may correspond to the same LR input. Many models have been proposed in the literature with multiple architectures with the aim of addressing this ambiguity using different architectural and optimization strategies. In our work we have tested four baseline models (Variational autoencoders \cite{chira2022imagesuperresolutiondeepvariational}; SRCNN \cite{srcnn}, SRGAN \cite{srgan}, SwinIR \cite{swinir}) used in comparative studies for super resolution problem \cite{ZHANG2025104995, 10874056, nikroo2023comparativeanalysissrganmodels, 10.1117/12.3032724}.

VAE-based SR \cite{chira2022imagesuperresolutiondeepvariational} formulates the problem in a probabilistic generative framework. A deep hierarchical VAE is trained to model the conditional distribution $p(\mathbf{x}_{HR} \mid \mathbf{x}_{LR})$ by introducing latent variables and optimizing the evidence lower bound (ELBO), which combines a reconstruction term with a Kullback–Leibler (KL) divergence regularization. In this formulation, the encoder maps the LR image to a latent representation, while the decoder generates HR samples conditioned on both the latent code and LR features. This allows the model to capture uncertainty and produce diverse high-frequency details rather than a single deterministic estimate. In contrast, SRCNN \cite{srcnn} is a deterministic convolutional neural network that directly learns an end-to-end mapping between LR and HR images. The LR image is first upsampled using bicubic interpolation to the desired scale, after which a three-layer CNN performs (i) patch extraction and representation, (ii) nonlinear mapping between LR and HR feature spaces, and (iii) reconstruction of the final HR output. The model is trained using mean squared error (MSE) loss, which encourages pixel-wise fidelity and typically leads to high PSNR performance, although it may produce overly smooth textures at large magnification factors. SRGAN \cite{srgan} extends this deterministic criterion by introducing adversarial learning to enhance perceptual quality. It consists of a deep residual generator network and a discriminator trained in a generative adversarial framework. Instead of relying solely on pixel-wise loss, SRGAN employs a perceptual loss composed of a content term (often computed in a high-level feature space such as VGG activations) and an adversarial term that pushes the generated images toward the natural image manifold. This design enables the synthesis of sharper and more realistic textures, especially at high upscaling factors, though it may not always maximize distortion-based metrics such as PSNR. Finally, SwinIR \cite{swinir} leverages Transformer-based attention mechanisms for image restoration. Its architecture consists of a shallow feature extraction layer, a deep feature extraction module built from Residual Swin Transformer Blocks (RSTBs), and a reconstruction module with sub-pixel convolution for upsampling. By employing shifted window-based self-attention, SwinIR captures both local and non-local dependencies efficiently while maintaining manageable computational complexity. Typically trained with an $\ell_1$ or pixel-wise loss for classical SR tasks, SwinIR balances distortion minimization and structural detail recovery, achieving strong performance across multiple restoration benchmarks.

We evaluate the four SR baselines at scales $\times2$, $\times4$, and $\times8$ under a unified training protocol. For each model and scale, we train for 20 epochs using Adam (learning rate $10^{-4}$, batch size $4$) on paired LR-HR samples. The split is deterministic (seed $42$) with 1656 training tiles and a held-out set of 414 tiles ($80/20$). SRCNN and the autoencoder are optimized with pixel-wise MSE after bicubic upsampling of LR inputs to HR size, SRGAN uses a weighted combination of MSE content loss and adversarial BCE loss, and SwinIR is trained with the same optimizer configuration and scale-specific LR-HR pairs. For each run, we retain the checkpoint with the best validation PSNR. Final evaluation is reported on the fixed SyMTRS test split ($n=414$) using MSE, PSNR expressed in Eq. \ref{eq:psnr}, and SSIM following \cite{ssim} expressed in Eq. \ref{eq:ssim}.

\begin{equation}
\mathrm{PSNR} = 10 \log_{10}\left(\frac{MAX_I^2}{\mathrm{MSE}}\right),
\label{eq:psnr}
\end{equation}
where $MAX_I$ is the maximum valid pixel value.

\begin{equation}
\mathrm{SSIM}(x,y)=\frac{(2\mu_x\mu_y+c_1)(2\sigma_{xy}+c_2)}{(\mu_x^2+\mu_y^2+c_1)(\sigma_x^2+\sigma_y^2+c_2)}.
\label{eq:ssim}
\end{equation}

\begin{figure}[ht]
    \centering
    \includegraphics[width=\linewidth]{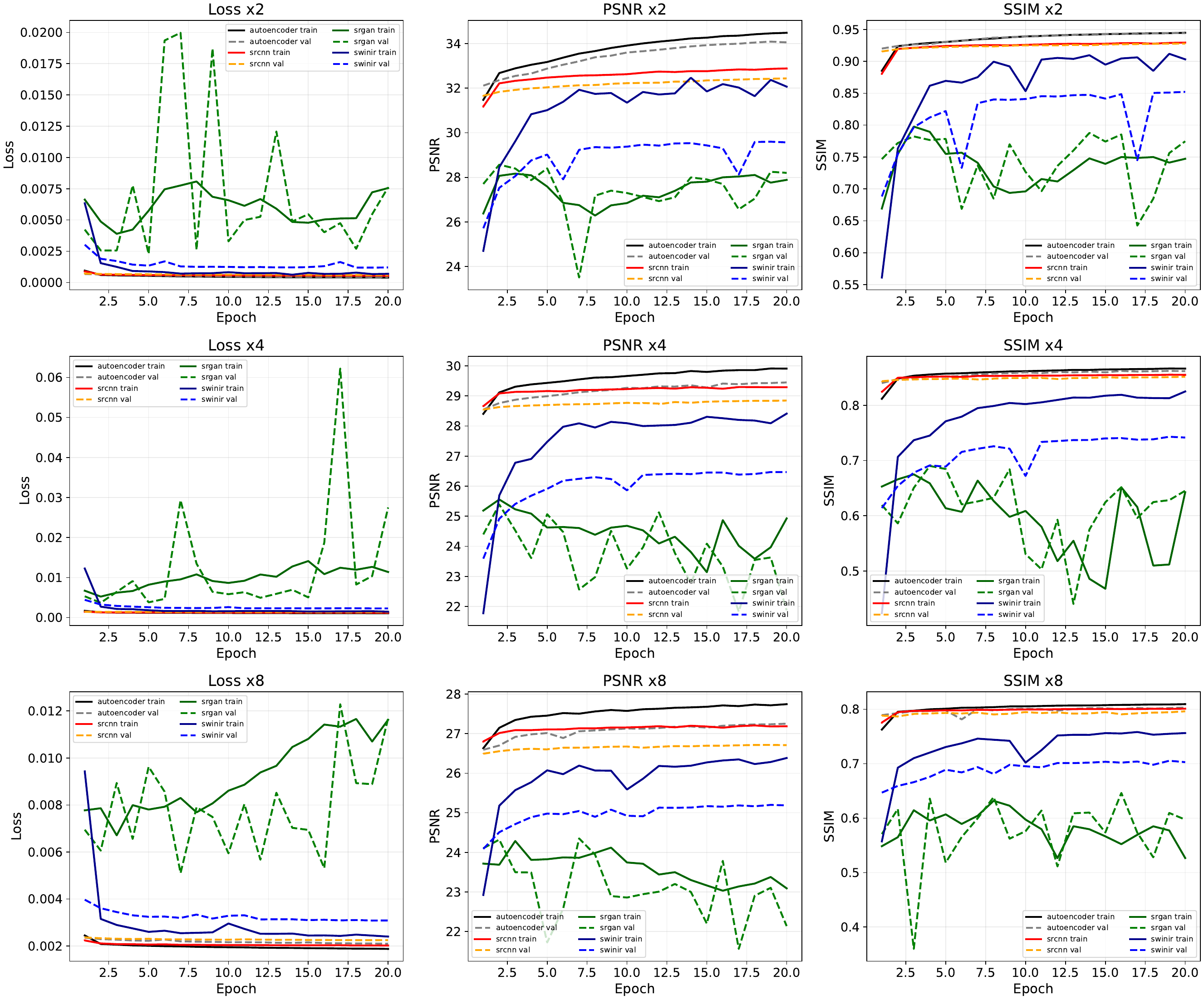}
    \caption{SR quantitative comparison aggregated from the test split for \(\times2\), \(\times4\), and \(\times8\).}
    \label{fig:training_curves}
\end{figure}

The training curves for all models and scaling factors are shown in Fig. \ref{fig:training_curves}. At $\times2$, all methods converge rapidly, with the autoencoder and SRCNN exhibiting smooth and stable optimization, minimal train--validation gaps, and the highest PSNR/SSIM values. SwinIR demonstrates a steady but slightly slower convergence, while SRGAN presents noticeable oscillations in both loss and validation metrics. As the scaling factor increases to $\times4$ and $\times8$, the super resolution problem becomes progressively more challenging, leading to lower PSNR and SSIM across all models. Nevertheless, the deterministic MSE-based approaches maintain stable convergence and consistent generalization. In contrast, SwinIR preserves relatively stable behavior with moderate performance degradation. Overall, the curves confirm the expected distortion--perception trade-off and highlight the increasing optimization difficulty as the magnification factor grows.

\subsection{Image-to-image translation}
One of the researched fields of generative AI is domain adaptation via image-to-image translation. Image-to-image translation aims to learn a mapping between two visual domains, either with paired supervision or from unpaired data. pix2pix \cite{isola2017pix2pix} formulates the problem in a fully supervised setting using conditional generative adversarial networks (cGANs). Given aligned training pairs $(x,y)$, the generator $G$ learns a direct mapping from input image $x$ to target image $y$, while the discriminator $D$ evaluates whether the generated output is indistinguishable from real samples conditioned on the same input. Architecturally, the generator follows a U-Net encoder–decoder structure with skip connections that transfer low-level spatial information directly from encoder to decoder layers, preserving fine details during reconstruction. The discriminator is implemented as a PatchGAN, which classifies local image patches instead of the entire image, encouraging high-frequency realism. Training optimizes a composite objective (Eq. \ref{eq:pix2pix}) combining an adversarial loss with an $\ell_1$ reconstruction term, 

\begin{equation}
\mathcal{L} = \mathcal{L}_{cGAN}(G,D) + \lambda \mathcal{L}_{\ell_1}(G),
\label{eq:pix2pix}
\end{equation}

where the $\ell_1$ loss enforces pixel-level fidelity to the ground-truth target while the adversarial component promotes perceptually realistic textures.

In contrast, CycleGAN \cite{zhu2017cyclegan} addresses the more challenging unpaired translation scenario, where aligned image pairs are unavailable. It learns two mappings, $G: X \rightarrow Y$ and $F: Y \rightarrow X$, together with corresponding discriminators for each domain. Since no paired supervision exists, CycleGAN introduces a cycle-consistency constraint that enforces $F(G(x)) \approx x$ and $G(F(y)) \approx y$, ensuring that translations remain structurally consistent with the input content. The overall objective combines adversarial losses for both domains with a cycle-consistency term and, optionally, an identity loss to preserve color composition when appropriate. Generators are typically implemented using residual convolutional networks with downsampling, residual blocks, and learned upsampling, while discriminators again employ PatchGAN structures. Unlike pix2pix, which directly minimizes pixel-wise discrepancy to a known target, CycleGAN primarily matches distributions across domains while preserving invertible structure through cycle consistency, making it suitable for style and appearance transfer when paired data is not available.

\section{Results and Discussion}
\subsection{Super resolution}
\begin{figure}[ht]
    \centering
    \includegraphics[width=\linewidth]{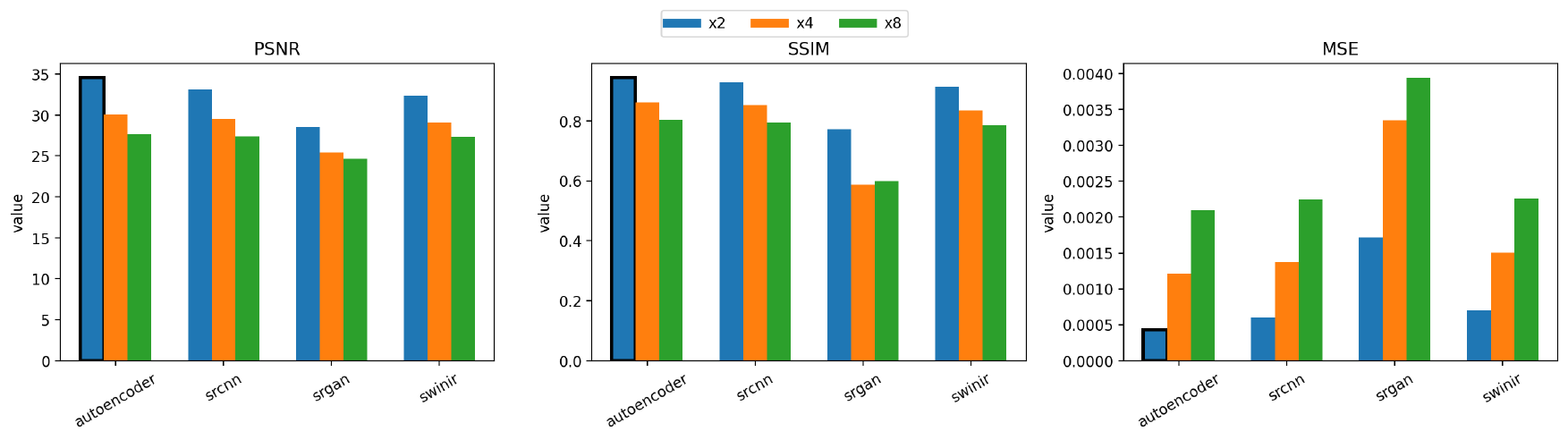}
    \caption{SR quantitative comparison aggregated from the test split for \(\times2\), \(\times4\), and \(\times8\).}
    \label{fig:sr_metrics_bar}
\end{figure}

Fig. \ref{fig:sr_metrics_bar} represent the histogram of rankings across all magnification factors for the super-resolution. The autoencoder achieves the best distortion-oriented performance at $\times2$, $\times4$, and $\times8$, obtaining the lowest MSE together with the highest PSNR and SSIM in every setting. At $\times2$, it reaches 34.564 dB PSNR and 0.9445 SSIM, outperforming SRCNN by 1.449 dB and 0.0167 SSIM, and SRGAN by almost 6 dB and 0.17 SSIM. The same tendency remains visible at $\times4$ and $\times8$, although the gap to the strongest deterministic baselines becomes smaller as the problem becomes harder. In particular, at $\times8$ the margin between the autoencoder, SRCNN, and SwinIR is below 0.32 dB PSNR, suggesting that once a large amount of high-frequency content has been removed by aggressive downsampling, all distortion-minimizing models approach a similar ceiling.
The scale-dependent degradation is also coherent with the expected difficulty of the benchmark. For the best-performing method, PSNR drops from 34.564 dB at $\times2$ to 30.023 dB at $\times4$ and 27.675 dB at $\times8$, while SSIM decreases from 0.9445 to 0.8622 and 0.8032 respectively. This decline indicates that the reconstruction task becomes progressively harder as the scale factor increases. Because the LR-HR pairs are perfectly aligned and generated from the same synthetic scene content, the observed performance drop can be attributed primarily to the loss of spatial detail rather than to registration noise or temporal mismatch, which often confound evaluation on real remote-sensing SR datasets.

\begin{figure}[ht]
    \centering
    \includegraphics[width=\linewidth]{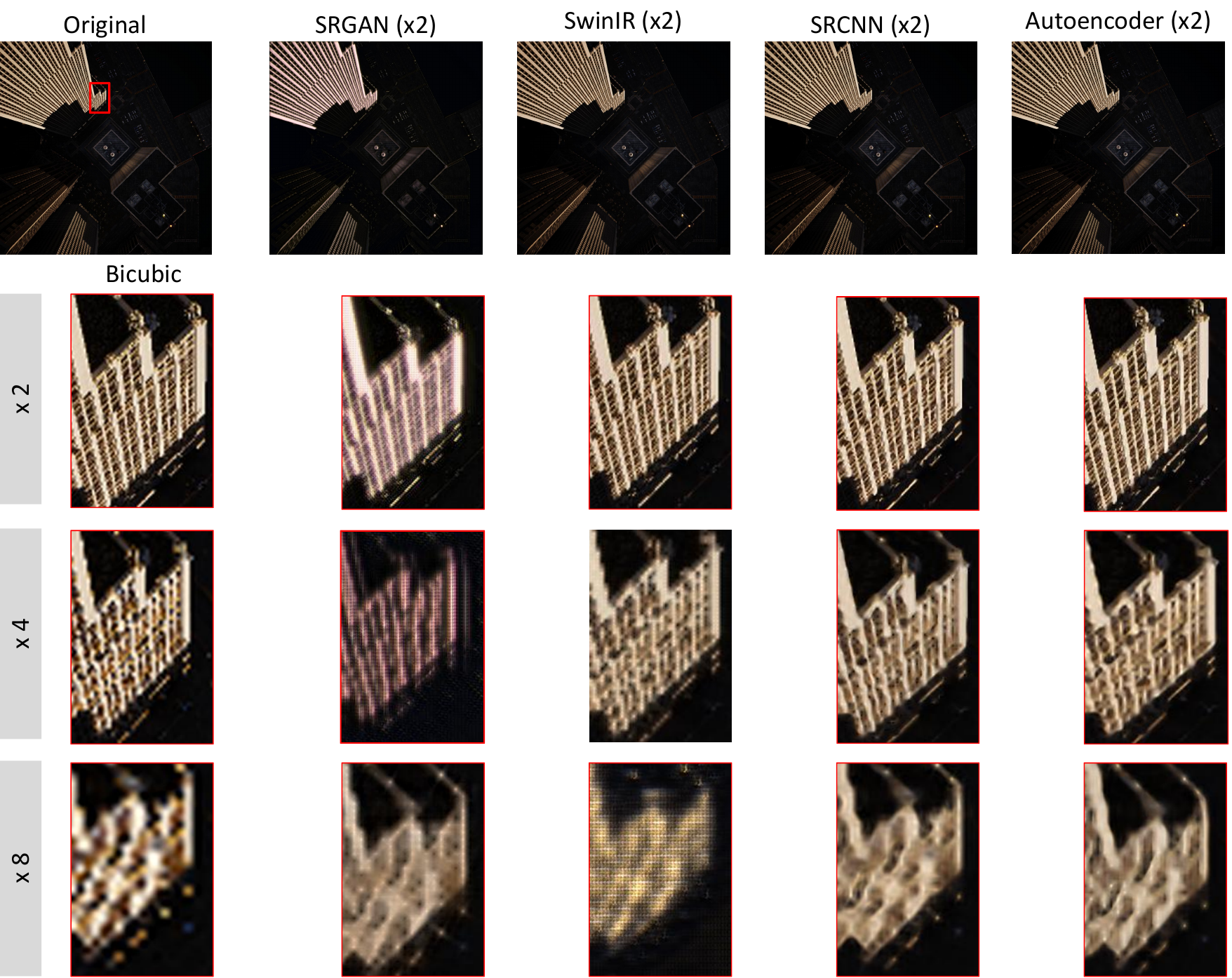}
    \caption{Qualitative SR comparison on examples degraded at scales \(\times2\), \(\times4\), and \(\times8\), reconstructed by the trained Autoencoder, SRGAN, SRCNN, and SwinIR models.}
    \label{fig:sr_visual_preview}
\end{figure}

Among the non-adversarial architectures, SRCNN remains highly competitive, especially at $\times4$ and $\times8$, where its results are very close to the autoencoder. This suggests that the benchmark is sufficiently structured for relatively shallow convolutional models to recover a substantial part of the missing information when optimized directly for pixel fidelity. SwinIR also delivers stable results, but under the present training budget it does not surpass the best convolutional baselines. A plausible interpretation is that the Transformer-based architecture requires either longer training, stronger hyperparameter tuning, or larger data diversity to fully exploit its modeling capacity. In other words, the current benchmark is already informative enough to separate architectures, while still leaving room for future gains from stronger optimization and model scaling.
SRGAN presents the weakest numerical performance at every scale, with the gap becoming particularly large in SSIM, even though adversarial training prioritizes perceptual sharpness and texture realism over exact pixel reconstruction. The qualitative examples in Fig. \ref{fig:sr_visual_preview} support this interpretation. The autoencoder and SRCNN generally preserve roof boundaries, road markings, and building outlines better, while SRGAN tends to generate visually sharper but less stable local patterns and stronger tonal deviations. SwinIR often recovers plausible structures, but still exhibits slightly softer or less consistent details than the top-performing autoencoder in these examples. 

% \begin{table*}[ht]
%     \centering
%     \caption{Super-resolution results on the SyMTRS test split (414 samples). Best values per scale are in bold.}
%     \label{tab:sr_results}
%     \begin{tabular}{c l c c c}
%         \toprule
%         \textbf{Scale} & \textbf{Model} & \textbf{MSE}$\downarrow$ & \textbf{PSNR (dB)}$\uparrow$ & \textbf{SSIM}$\uparrow$ \\
%         \midrule
%         $\times2$ & Autoencoder & \textbf{0.000435} & \textbf{34.564} & \textbf{0.9445} \\
%         $\times2$ & SRCNN & 0.000598 & 33.115 & 0.9278 \\
%         $\times2$ & SRGAN & 0.001719 & 28.569 & 0.7720 \\
%         $\times2$ & SwinIR & 0.000700 & 32.379 & 0.9135 \\
%         \midrule
%         $\times4$ & Autoencoder & \textbf{0.001213} & \textbf{30.023} & \textbf{0.8622} \\
%         $\times4$ & SRCNN & 0.001371 & 29.513 & 0.8518 \\
%         $\times4$ & SRGAN & 0.003349 & 25.390 & 0.5862 \\
%         $\times4$ & SwinIR & 0.001507 & 29.091 & 0.8353 \\
%         \midrule
%         $\times8$ & Autoencoder & \textbf{0.002097} & \textbf{27.675} & \textbf{0.8032} \\
%         $\times8$ & SRCNN & 0.002247 & 27.384 & 0.7945 \\
%         $\times8$ & SRGAN & 0.003945 & 24.621 & 0.5998 \\
%         $\times8$ & SwinIR & 0.002263 & 27.356 & 0.7864 \\
%         \bottomrule
%     \end{tabular}
% \end{table*}

\subsection{Image-to-image translation}
\begin{figure}[ht]
    \centering
    \includegraphics[width=\linewidth]{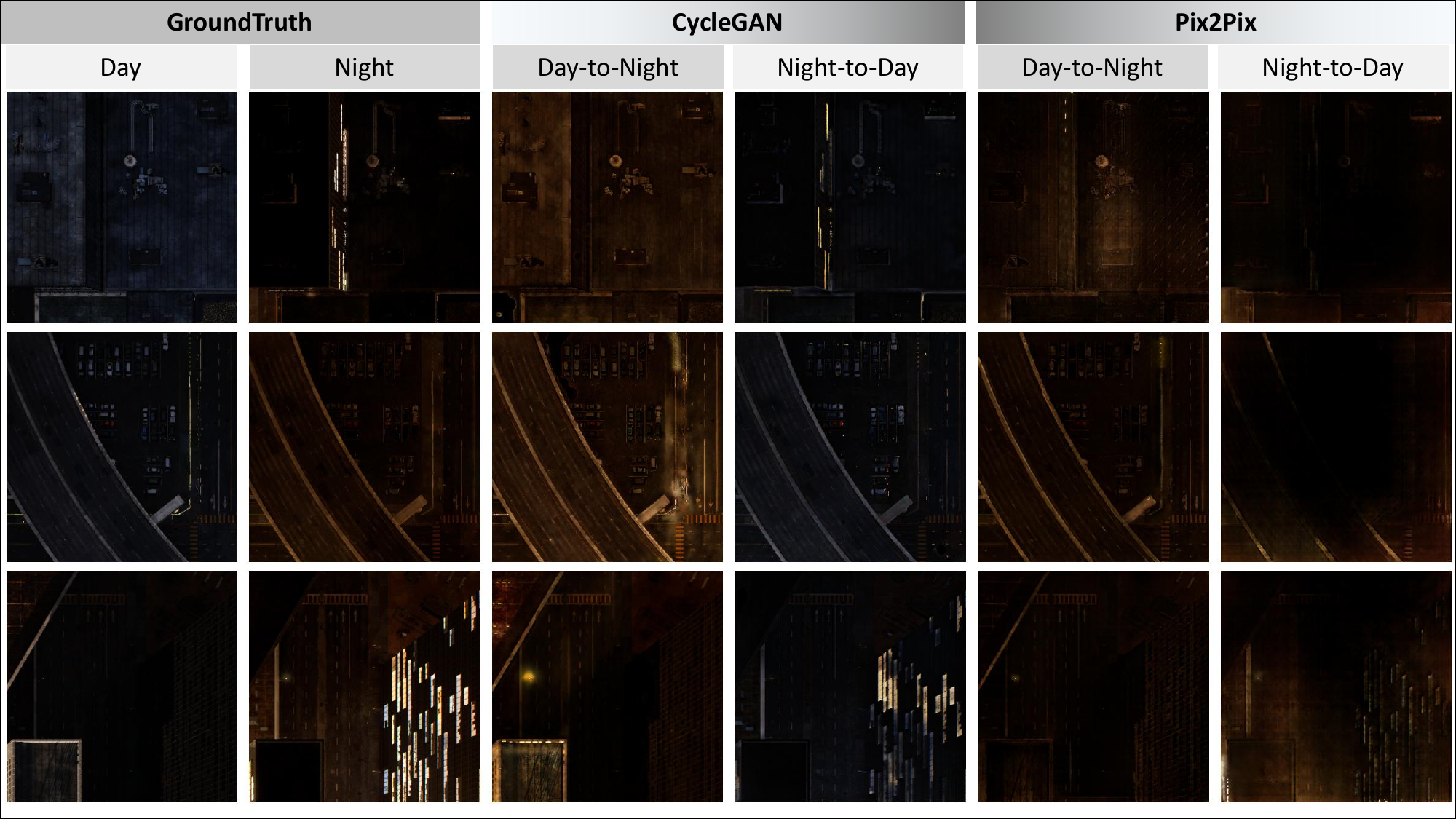}
    \caption{Qualitative comparison for day-to-night and night-to-day translation using CycleGAN and pix2pix on SyMTRS examples.}
    \label{fig:Im2Im_translation}
\end{figure}

Figure \ref{fig:Im2Im_translation} illustrates representative day-to-night translations obtained with CycleGAN and pix2pix. A first result is that both models preserve the global scene geometry well including road layout, building footprints, lane markings, and bright window patterns remain spatially aligned with the input views. 
The visual comparison also highlights a difference between paired and unpaired training. In the day-to-night direction, pix2pix generally produces outputs that are closer to the target night domain, with darker global illumination, more localized light sources, and better preservation of scene-specific contrast transitions. CycleGAN, by contrast, often applies a more global dark surface and tends to brighten some surfaces excessively, especially roads and rooftops.
The night-to-day direction appears more challenging for both methods, due to nighttime containing less visible information in dark regions. In the examples shown, CycleGAN often recovers daytime structure more clearly, producing a more readable road network and facade layout, albeit sometimes with a cooler tone than the ground truth. Pix2pix, despite the advantage of paired supervision, occasionally yields underexposed outputs with limited recovery in shadowed areas. This asymmetry suggests that translating from information-poor nighttime observations back to daytime appearance remains a difficult inverse problem even in a controlled synthetic setting.

Therefore, SyMTRS can support both paired and unpaired translation modes while keeping the scene content consistent enough for meaningful visual comparison. Additionally, they indicate that the dataset is challenging enough to expose model-specific failure modes: CycleGAN may sacrifice sample-specific fidelity for style consistency, whereas pix2pix can better exploit paired alignment but may still struggle when the source image lacks too much texture information.

\newpage

\section*{Code availability}
The dataset generated via Unreal Engine 5 and used to train the models for this paper is available at \textbf{HuggingFace} \url{https://huggingface.co/datasets/safouaneelg/SyMTRS}. The code and weights are available in the \textbf{Github} repository \url{https://github.com/safouaneelg/SyMTRS}.

\section{Conclusion}
In this paper, we introduced \textbf{SyMTRS}, a synthetic multi-task remote sensing dataset designed to enable controlled research in aerial image super-resolution and day/night image translation through perfectly aligned high-resolution imagery, multi-scale low-resolution counterparts, and paired cross-domain samples generated within a unified simulation pipeline. The reported benchmarks show that SyMTRS is both reliable and sufficiently challenging: super-resolution performance degrades consistently from $\times2$ to $\times8$, confirming the increasing difficulty of high-magnification reconstruction, while the compared baselines exhibit clear and stable differences, with the autoencoder achieving the strongest distortion-based performance and adversarial translation models exposing the expected trade-offs between perceptual realism and pixel fidelity. The image-to-image translation results demonstrate that the dataset preserves scene geometry across domains and can support both paired and unpaired adaptation studies.

\bibliographystyle{elsarticle-harv}
\bibliography{references}
\end{document}